\def\year{2020}\relax
\documentclass[letterpaper]{article} 
\usepackage{aaai20}  
\usepackage{times}  
\usepackage{helvet} 
\usepackage{courier}  
\usepackage[hyphens]{url}  
\usepackage{graphicx} 
\usepackage{graphicx}  
\usepackage{tabularx}
\usepackage{threeparttable}
\usepackage{multirow}
\usepackage{booktabs}
\usepackage{amssymb}
\usepackage{amsmath}
\usepackage{tcolorbox}
\usepackage{footnote}
\newcommand{\citet}[1]{\citeauthor{#1} \shortcite{#1}}
\newcommand{\citep}{\cite}

\DeclareMathOperator*{\argmax}{arg\,max}

\title{A Pre-training Based Personalized Dialogue Generation Model with Persona-sparse Data}
\author{
Yinhe Zheng \textsuperscript{\rm 1,3}\thanks{Equal contributions},
Rongsheng Zhang\textsuperscript{\rm 2}\footnotemark[1],
Xiaoxi Mao\textsuperscript{\rm 2},
Minlie Huang\textsuperscript{\rm 1}\thanks{Corresponding Author: Minlie Huang}\\
\textsuperscript{\rm 1} Institute for Artifical Intelligence, State Key Lab of Intelligent Technology and Systems. \\
Beijing National Research Center for Information Science and Technology.\\
Department of Computer Science and Technology, Tsinghua University, Beijing, China. \\
\textsuperscript{\rm 2} Fuxi AI Lab, NetEase Inc., Hangzhou, China. \\
\textsuperscript{\rm 3} Samsung Research China - Beijing (SRC-B), Beijing, China. \\
yh.zheng@samsung.com, zhangrongsheng@corp.netease.com, maoxiaoxi@corp.netease.com, \\ aihuang@tsinghua.edu.cn
}
\begin{document}

\maketitle

\begin{abstract}
Endowing dialogue systems with personas is essential to deliver more human-like conversations. However, this problem is still far from well explored due to the difficulties of both embodying personalities in natural languages and the persona sparsity issue observed in most dialogue corpora. This paper proposes a pre-training based personalized dialogue model that can generate coherent responses using persona-sparse dialogue data. In this method, a pre-trained language model is used to initialize an encoder and decoder, and personal attribute embeddings are devised to model richer dialogue contexts by encoding speakers' personas together with dialogue histories. Further, to incorporate the target persona in the decoding process and to balance its contribution, an \emph{attention routing} structure is devised in the decoder to merge features extracted from the target persona and dialogue contexts using dynamically predicted weights. Our model can utilize persona-sparse dialogues in a unified manner during the training process, and can also control the amount of persona-related features to exhibit during the inference process. Both automatic and manual evaluation demonstrates that the proposed model outperforms state-of-the-art methods for generating more coherent and persona consistent responses with persona-sparse data.
\end{abstract}

\section{Introduction}\label{sec:intro}
Building a ``human-like'' conversation system has been an important topic in artificial intelligence, where one of the major challenges is to present a consistent persona so that the system can interact with users in a more natural way to gain users' confidence and trust. The user engagement of a dialogue agent increases when the agent is conditioned on various persona settings, including age, gender, language, location, or even a proper accent~\citep{shum2018eliza,wang2018chat,huang2019challenges,Zhou2018Commonsense}. Various approaches have been explored to personalize a dialogue system~\citep{Li2016_ACL,Qian2018Assigning,Kottur2017Exploring}. 

\begin{figure}[htb!]
\centering
\includegraphics[width=180px]{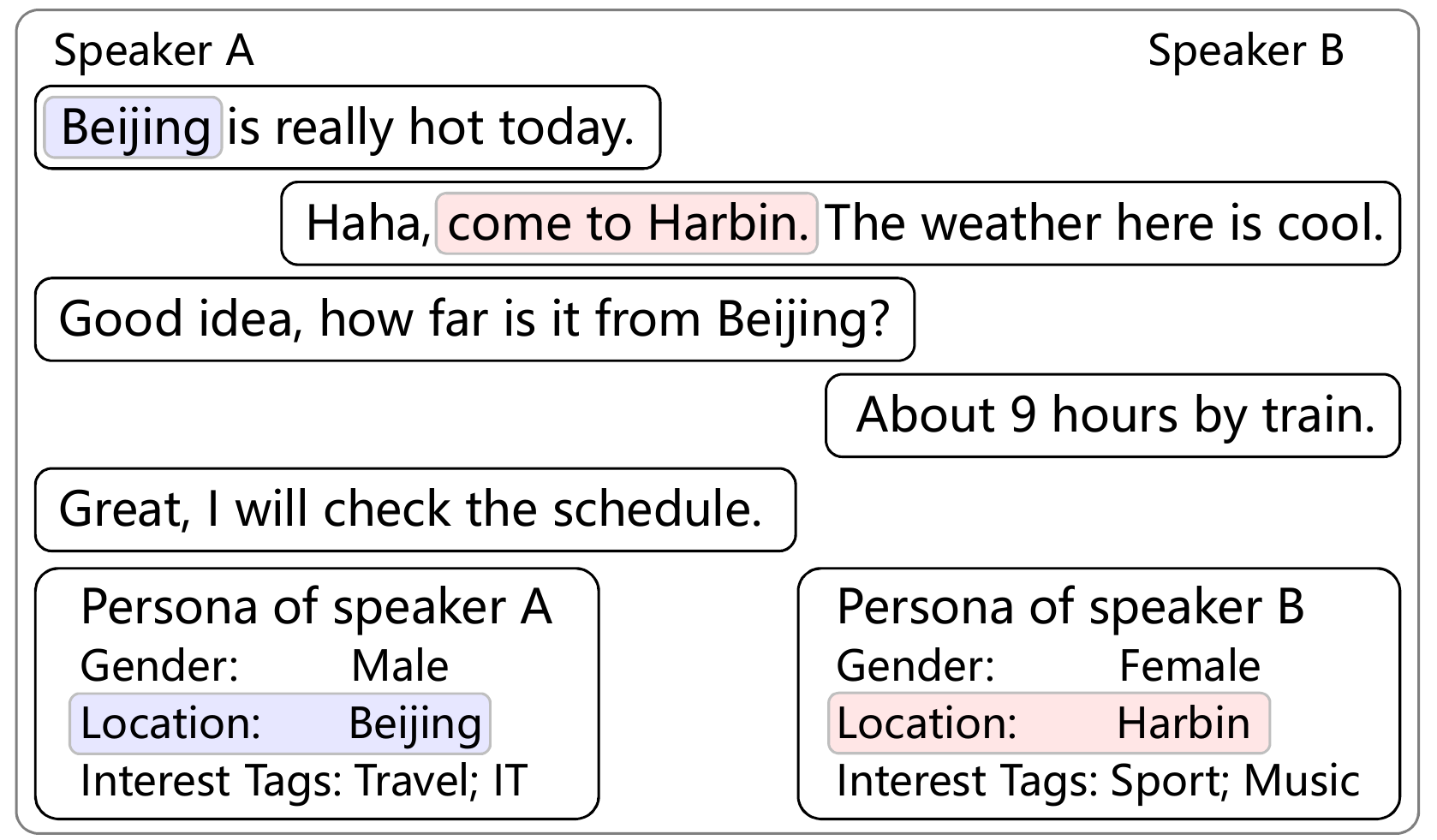}
\caption{An example dialogue session and the personal profile of each speaker. Words in responses are in the same color with the corresponding personal attributes.}
\label{fig:example}
\end{figure} 

Recent advances in pre-training methods have led to state-of-the-art results in a range of natural language processing tasks~\citep{peters-etal-2018-deep,devlin2018bert,radford2019language,ke2019sentilr}. Promising results are also obtained by applying these approaches in personalized dialogue generation models, such as fine-tuning a pre-trained model on a small set of persona-related dialogues (e.g. PERSONA-CHAT~\citep{Zhang2018Personalizing})~\citep{mazare2018training,wolf2018transfer,golovanov-etal-2019-large}. However, the dialogue data used in the fine-tuning stage of these methods are usually crowd-sourced, where speakers are required to exchange their personas within limited turns of conversation. This data collection scheme is guaranteed to yield dialogues that cover rich persona related features (i.e., ``\textbf{persona-dense}'') and thus facilitate fine-tuning directly. However, such a setting is expensive and can only produce a limited amount of dialogues. Further, models fine-tuned using these data tend to over-fit to the routine that persona-related features should be exhibited in every response. This does not meet the common practice observed in our daily communication.

As a matter of fact, most speakers in our daily conversations are not aiming to exhibit their personas within limited turns of interactions, namely, real-world dialogues are not always persona-related. For example, as shown in the dialogue session of Figure~\ref{fig:example}, speaker A and B only reveal their personas in the first turn of the conversation, while the rest part of this dialogue is not persona-related. Therefore, we argue that data collected from real-world conversations only contain a limited amount of dialogues that relate to speakers' persona. In other words, real dialogue data are ``\textbf{persona-sparse}''. Directly fine-tuning on these real-world conversations may mislead the model to focus on dialogues that are not persona-related, since these dialogues are in the majority. Further, speakers' personas may be regarded as the noises and tend to be ignored by the dialogue model since they are not related to most responses.

To address the above issues, we propose a pre-training based method that can utilize persona-sparse data to build a personalized dialogue agent. Specifically, the dialogue data we use come from the daily conversations on social media, where speakers are not asked to reveal their personas \emph{intentionally}. This differs from previous pre-training based approaches that utilize crowdsourced dialog data such as PERSONA-CHAT~\citep{Zhang2018Personalizing}, which is persona-dense and thus a direct fine-tuning process will suffice for the pre-trained dialogue model to capture persona related features~\citep{wolf2018transfer,golovanov-etal-2019-large}.

In this paper, we adopt the encoder-decoder framework and use a pre-trained language model to initialize an encoder and decoder. Attribute embeddings are added in the encoder to capture rich persona related features when modeling dialogue histories, and an \emph{attention routing} mechanism is proposed in the decoder to incorporate the target persona in the decoding process. Three attention routes are used in this study and each route models a certain source of features, i.e., features extracted from the target persona, dialogue histories, and previously decoded tokens. A dynamic weight predictor is built to weigh the output of each route, so that the contribution of the target persona in the final output can be balanced. In this manner, we can leverage persona-sparse dialogue data in the training stage and control the amount of persona information to exhibit in the inference stage. Automatic and manual evaluation indicates that our method can produce dialogue responses that are more coherent and contain richer persona features.

Our main contributions can be summarized as follows:
\begin{enumerate}
    \item We propose a pre-training based method that can utilize persona-sparse data to build personalized dialogue models. Our method can take full advantage of the pre-trained model for generating diverse and coherent dialogues, while effectively leveraging real-world data that are persona-sparse to produce persona-related responses.
    
    \item We propose an attention routing mechanism to weigh persona features dynamically in the decoder. It allows us to utilize persona-sparse dialogue data in a unified manner during the training process and to control the amount of persona-related features to exhibit in the decoded responses.
    
    \item Both automatic and manual evaluation shows that our method outperforms previous methods in producing more coherent and persona-related responses.
\end{enumerate}

\section{Related Work}
\textbf{Personalized Dialogue Models:} Traditional studies to build personalized dialogue agents focused on psychology inspired approaches, such as modeling ``Big Five'' of speakers~\citep{mairesse2007personage}. However, such psychological metrics are too subjective to model and the corresponding dialogue data are extremely difficult to collect. This limits the application of these approaches in building large-scale personalized dialogue systems.

Recent studies try to tackle the personalized dialogue generation problem in a data-driven manner, i.e., learning persona related features directly from large-scale dialogue datasets. Early attempts in this direction focused on modeling characters in movie dialogues~\citep{Banchs2012Movie}, while recent studies took advantages of the sequence to sequence learning framework~\citep{sutskever2014sequence,serban2016building} to model a speaker's persona by utilizing social media data~\citep{zheng2019Personal}. Specifically, persona in these studies was either implicitly modeled using a persona embedding~\citep{Li2016_ACL,Kottur2017Exploring,luan2017multi} which requires sufficient data from each speaker, or explicitly given as the input~\citep{Qian2018Assigning,song2019exploiting}. Some models were also proposed to personalize task-oriented dialogue systems~\citep{Luo2019Personalized}. However, these models were all trained from scratch without a pre-training process, where the benefits of using language models that are pre-trained with large corpora are yet to be explored.

\textbf{Pre-training Methods:} Recent advances in the pre-training methods have led to state-of-the-art results in many tasks~\citep{peters-etal-2018-deep,radford2018improving,devlin2018bert,sun2019ernie}. Various pre-training approaches have also been proposed in the dialogue modeling task~\citep{zhang2017neural}. Specifically, \citet{mehri2019pretraining} proposed two pre-training objectives to boost dialogue tasks; \citet{budzianowski2019hello} adapted the pretrained GPT2 model~\citep{radford2019language} to multi-domain task-oriented dialogues. As for personalized dialogue modeling, \citet{wolf2018transfer} and \citet{golovanov-etal-2019-large} showed that the pre-trained GPT model~\citep{radford2018improving} can significantly improve the quality of the generated dialogues by fine-tuning on a small persona-dense dialogue dataset.

Compared to ours, the most relevant prior work was done by~\citet{golovanov-etal-2019-large}. However, their method requires a direct fine-tuning process on a persona-dense dataset, while our study can make use of the persona-sparse dialogues with the proposed dynamic weighting scheme. Further, we also add explicit attribute embeddings to model structured personas when encoding dialogue contexts, whereas their approaches do not consider speakers' personas when modeling dialogue contexts.

\section{Model}
\begin{figure*}[t]
    \centering
    \includegraphics[width=430px]{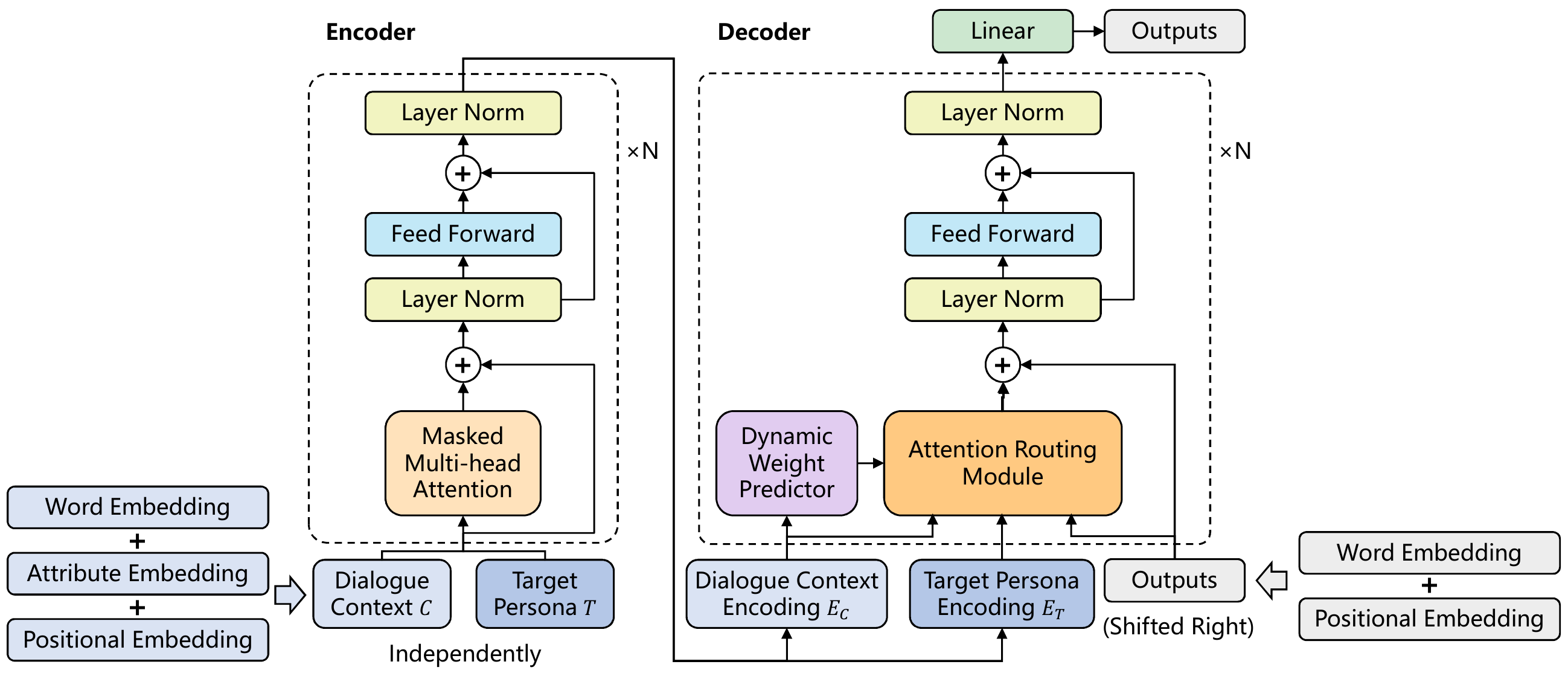}
    \caption{The framework of the proposed personalized dialogue generation model. The encoder and decoder share the same set of parameters. The dialogue context and the target persona are encoded independently using the encoder and their encodings are fed into the attention routing module in each decoder block. A dynamic weight predictor is trained to weigh the contribution of each route.}
    \label{fig:model_arch}
\end{figure*}

Our study aims at generating a fluent response $Y$ that is coherent with a given dialogue context $C$ and an explicitly represented target persona $T$ of the responder:
\begin{equation}
    Y = \argmax_{Y'} P({Y'}|C, T)
\end{equation}
Specifically, the persona $T$ can be regarded as a set of attributes (such as gender, location, or personal interest) $T$=$\{t_1, t_2, ..., t_N\}$ and each attribute is represented as a key-value pair $t_i$=$\langle k_i, v_i\rangle$. The dialogue context $C$=$\{(U_1, T_1), ..., (U_M, T_M)\}$ contains several turns of conversations (i.e., utterances $U_i$) and the persona $T_i$ of the associated speaker.

Figure~\ref{fig:model_arch} shows an overview of our model. The encoder and decoder used in our study follow the Transformer framework~\citep{Vaswani2017Attention}, and share the same set of weights. The encoder is used to encode the dialogue context $C$ into a context encoding $E_C$ and the target persona $T$ into a persona encoding $E_T$, independently. Attribute embeddings are added when producing $E_C$. The decoder takes as input $E_C$ and $E_T$ and decodes the output in an auto-regressive way. An attention routing mechanism is proposed by extending the original multi-head attention module and introducing a dynamic weight predictor. Outputs of these attention routes are merged using the dynamically predicted weight.

\subsection{Encoding with Personas}\label{sec:att_embed}
\begin{figure}[t]
    \centering
    \includegraphics[width=180px]{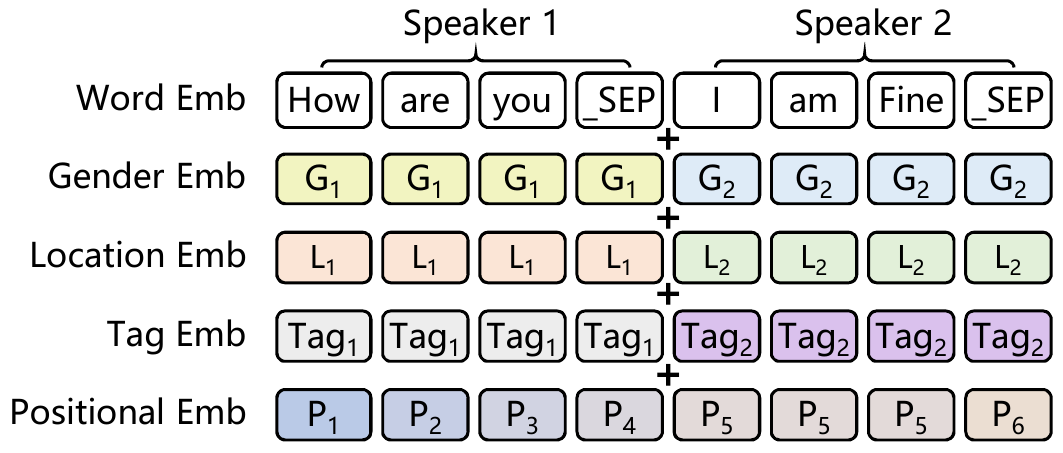}
    \caption{Input representation of the dialogue context. The input embedding for each token is the sum of a word embedding, a positional embedding, and attribute embeddings. Three kinds of attribute embeddings are modeled, i.e., gender embedding, location embedding, and tag embedding. The tag embedding of a speaker is calculated as the average of all the tag representations since each speaker may have several interest tags.}
    \label{fig:context_rep}
\end{figure}

We introduce attribute embeddings in the encoder to model each persona $T_i$, $(i = 1, 2, ..., n)$ that is involved in the dialogue context $C$. Specifically, we first concatenate all the utterances in $C$ with a special token ``\_SPE'' and map each attribute $t_j$ in $T_i$ to an embedding representation. The input embedding for the Transformer encoder at each time step is constructed by adding the word embedding, positional embedding and attribute embeddings together (Figure~\ref{fig:context_rep}). The proposed attribute embeddings enhance the dialogue context encoding $E_C$ since the persona of every speaker is modeled in $E_C$. This differs from the previous work of~\citet{golovanov-etal-2019-large}, where only word embeddings are used.

More precisely, three attributes are modeled in this study: Gender, Location, and Interest Tags. The embedding of Gender and Location can be obtained simply utilizing look-up tables since these attributes only have one unique value for each speaker, while the embedding of Interest Tags is computed as the average of all the tag embeddings for a speaker since each speaker may have several different interest tags.

Moreover, for the target persona $T$ that the generated response should be coherent to, we pack all the key-value pairs in $T$ into a word sequence and feed the corresponding word embeddings to the encoder to obtain $E_T$.

\subsection{Attention Routing}
In order to effectively utilize the persona-sparse dialogue data in a unified manner, it is expected to involve little or no persona features in the decoding process when training on non-persona-related dialogues, whereas to incorporate abundant persona features when modeling persona-related dialogues. In this study, we devise an attention routing mechanism in the decoder to control the contribution of the target persona $E_T$ in the decoding process.

Specifically, the vanilla multi-head attention module in the original Transformer block is extended to model the encodings of the target persona $E_T$, the dialogue context $E_C$ and previously decoded tokens $E_{prev}$, respectively. We name each set of the multi-head attention operation as an \emph{attention route} since it routes to different input features. More specifically, The three attention routes in our study use features extracted from previously decoded tokens $E_{prev}$ as the query, and attend to $E_T$, $E_C$ and $E_{prev}$, respectively, i.e.,
\begin{align}
 O_T &= {\rm MultiHead}(E_{prev}, E_T, E_T)\label{eq:persona_enc} \\
 O_C &= {\rm MultiHead}(E_{prev}, E_C, E_C)\label{eq:context_enc} \\
 O_{prev} &= {\rm MultiHead}(E_{prev}, E_{prev}, E_{prev})\label{eq:prev_enc}
\end{align}
Here, we use unmasked bi-directional self-attention in Eq.\ref{eq:persona_enc} and \ref{eq:context_enc} to facilitate more efficient interactions, and use masked self-attention in Eq.~\ref{eq:prev_enc} to avoid feeding the ``golden truth'' token.

The outputs of each attention route $O_T$, $O_C$ and $O_{prev}$ are averaged using a persona weight $\alpha \in [0, 1]$:
\begin{equation}\label{eq:mean_att}
    O_{merge} = \alpha O_T + (1- \alpha)O_C + O_C + O_{prev}
\end{equation}
where a large $\alpha$ value indicates that more persona information will flow to the final outputs, and thus the generated responses are expected to exhibit more persona-related features. Note that Eq.~\ref{eq:mean_att} ensures that features extracted from the dialogue context $O_C$ and previous decoded tokens $O_{prev}$ can always be incorporated in the decoder.

Ideally, the value of $\alpha$ should be annotated based on whether the training dialogue is persona-related or not. However, this would be impractical for a large scale dialogue dataset. In this study, we design a dynamic weight predictor to calculate $\alpha$ automatically in the training stage. Specifically, the predictor is modeled as a binary classifier $P_\theta(r|E_c)$ that takes as input the dialogue context encoding $E_C$ and predicts whether the training dialogue is persona related ($r=1$) or not ($r=0$). The confidence of this binary classifier is used as the predicted weight:
\begin{equation}
    \alpha = P_\theta(r=1| E_C)
\end{equation}

More precisely, we model the weight predictor using a neural network that is parameterized by $\theta$, and develop a heuristic script to produce labels for the training dialogue to optimize $\theta$. This script applies manually crafted rules such as word matching to decide whether a given dialogue is persona-related or not. The weight predictor is jointly trained with the dialogue model by optimizing the following cross entropy loss on these script-generated noisy labels:
\begin{equation*}
    L_{W}(\theta) = -\sum_i r_i{\rm log}P_\theta(r_i | E_C) + (1 - r_i){\rm log}[1 - P_\theta(r_i | E_C)]
\end{equation*}

Note that we can also directly use the heuristic script as the weight predictor, i.e., set $\alpha$ to either 1 (the input dialogue is persona-related) or 0 (otherwise) in the training process. However, these hard decisions may bring bias introduced by the script to our model and thus lead to sub-optimal results. On the contrary, our neural-based predictor models a soft approximation of the prior knowledge provided by the heuristic script, and the joint training approach also guide the encoder to capture more persona related features in the context encoding $E_C$. Our experiments also verify that the soft weights produced by our predictor lead to better results compared to the hard weights produced by the heuristic script.

\subsection{Pre-training and Fine-tuning}
A pre-training process is employed in this study. Specifically, we collect a large set of text data and follow the scheme introduced by~\citep{radford2018improving} to train a language model by optimizing the standard maximum log likelihood loss:
\begin{equation}\label{eq:lm_loss}
    L_{LM}(\phi) = -\sum_i {\rm log}P_\phi(u_i | u_{i-k}, \dots, u_{i-1})
\end{equation}
where $\phi$ is the parameter set of the language model, $k$ is the size of the context window, and $u_{i-k}, \dots, u_{i-1}, u_i$ is a sequence of tokens that is sampled from the training corpus.

Once pretrained, the parameter set $\phi$ is used to initialize the encoder and decoder of our model, and a fine-tuning process is employed to adapt $\phi$ to the dialogue dataset. We optimize the following loss for the dialogue generation task:
\begin{equation}
    L_{D}(\phi) = -\sum_i {\rm log}P_\phi(u_i | u_{i-k}, \dots, u_{i-1}, E_C, E_T)
\end{equation}
where $E_C$ and $E_T$ is the dialogue context and target persona encoding, respectively, and $u_{i-k}, \dots, u_{i-1}, u_i$ is a sequence of tokens from the dialogue response.

Further, in order to bridge the gap between the data used in the pre-training and the fine-tuning stage, we also optimize the language model loss (i.e., Eq.~\ref{eq:lm_loss}) that is evaluated using utterances sampled from the dialogue contexts in the fine-tuning process. This is in line with the prior work~\citep{radford2018improving}, in which performance improvements are observed when incorporating such an auxiliary loss.
Specifically, $L_{LM}(\phi)$ is optimized in the pre-training stage and the following loss is optimized in the fine-tuning stage:
\begin{equation}\label{eq:total_loss}
    L(\phi, \theta) = L_{D}(\phi) + \lambda_1 L_{LM}(\phi) + \lambda_2 L_{W}(\theta)
\end{equation}
where $\lambda_1$ and $\lambda_2$ are hyper-parameters to balance each loss.

\section{Dataset}
The dialogue data used in this study were sampled from the PersonalDialog dataset~\citep{zheng2019Personal}, which were collected from a Chinese social media Weibo. Each dialogue in this dataset is composed of a Weibo post and its following replies. A structured personal profile of each speaker was also provided in this dataset, and three persona attributes (i.e., ``Gender'', ``Location'' and ``Interest Tags'') were approached in our study. Figure~\ref{fig:example} shows a sampled dialogue and Table~\ref{tab:dataset} shows a basic statistic of the data used in this study. About 0.88M dialogues are labeled as persona-related by our heuristic script.

\begin{table}[tb]
\centering
\caption{Statistics of the dialogue dataset used in this study.}
 \begin{tabular}{ll}
  \toprule
  Total number of dialogues             & 5.44 M  \\
  Total number of speakers              & 1.31 M  \\
  Total number of utterances            & 14.40 M \\
  Dialogues with more than 4 utterances & 0.81 M  \\
  Average utterances per dialogue       & 2.65    \\
  Average tokens per utterance          & 9.46    \\
  \bottomrule
 \end{tabular}
\label{tab:dataset}
\end{table}

We randomly sampled 10K sessions of dialogues as the validation set, and constructed two test sets, i.e., a random test set and a biased test set, to test the behavior of our model in different contexts. Specifically, the random test set contained 10K sessions of dialogues that were randomly sampled. Most of these dialogues did not contain persona-related features since common Weibo users are not required to exhibit their personas intentionally on Weibo. Therefore, the contexts provided by the random test set are persona-sparse.

The biased test set was deliberately chosen to provide us different contexts under which speakers tend to reveal their personas. For example, the dialogue ``Are you a boy or a girl?'' and ``I am a girl'' is biased since the speaker reveals her gender in response to the gender-related post. We manually labeled 521 biased dialogues in this study. The contexts provided by the biased test set are persona-dense. It will be interesting to see if our model can produce more persona consistent responses under these biased contexts.

\section{Experiments}\label{sec:exp}
\subsection{Persona Classifier}\label{sec:persona_cls}
In order to better evaluate whether the generated dialogue responses carry rich persona-related features, we built a binary classifier that takes as input a dialogue response $R$ and a persona $T$ in the form of a concatenated token sequence, and predicts whether $T$ is exhibited in $R$. Specifically, we randomly sampled a batch of response-persona pairs, and manually labeled 1,044 positive pairs such that the persona is exhibited in the response. We also labeled the same number of negative pairs such that the response do not carry persona related features. We split these data into the train, validation, and test set with the ratio of 8:1:1 and fine-tuned a classifier based on the BERT-base model~\citep{devlin2018bert}. The accuracy of this classifier on the test set achieved 75.5\%. Similar approach was also used by \citet{zhou2018emotional} and \citet{zheng2019Personal}.

\subsection{Implementation Details}
Our pre-training stage used a dataset that was collected from a set of Chinese novels, which covered a variety of genres (including Comedy, Romance, Mystery). The final pretraining corpus contains about 0.5 billion tokens and we trained a character-level language model with a vocabulary size of 13,084. The encoder and decoder contained 12 Transformer blocks, and 12 attention heads were used. Token embeddings had the size of 768 and the context window was of size 512. The dynamic weight predictor was implemented as a multi-layer perceptron after an average pooling layer on $E_C$. The value of $\lambda_1$ and $\lambda_2$ in Eq.~\ref{eq:total_loss} was set to 0.2 and 0.5, respectively. The pretraining stage lasted for 70 epochs, and we fine-tuned our model for another 30 epochs.

\subsection{Baselines}
We chose several baselines:
\begin{itemize}
  \item \textbf{Att+PAB}: A RNN based model that encodes the input persona into a representation vector using a persona fusion module, and decodes personalized responses utilizing a persona-aware bias~\citep{zheng2019Personal}.
  
  \item \textbf{Trans.}: A Transformer model~\citep{Vaswani2017Attention} that takes concatenated dialogue histories as input and produces the corresponding dialogue response without using persona-related features.
  
  \item \textbf{TTransfo}: The TransferTransfo model as introduced by~\citet{wolf2018transfer}. This model fine-tunes the pre-trained model directly on the persona-sparse dialogues.
  
  \item \textbf{TTransfo + P}: The same as the TransferTransfo model but this model is fine-tuned using only dialogues that are labeled as persona-related by our heuristic script, i.e., noisy persona dense dialogue data.
  
  \item \textbf{LConv}: The \emph{multi-input} model~\footnote{This model was proposed by the winning team ``Lost in Conversation'' in the ConvAI2 competition~\citep{dinan2019convai2}.} proposed in ~\citep{golovanov-etal-2019-large}. This model uses a pre-trained encoder and decoder and is fine-tuned directly on the persona-sparse dialogues without using the dynamic weight predictor.
  
  \item \textbf{LConv + P}: The same as the \emph{multi-input} model but it is fine-tuned using only dialogues that are labeled as persona-related by our heuristic script.
\end{itemize}
All the baselines that utilize the Transformer architecture used the same set of hyper-parameters, and the same pre-training approach is employed.

Further, we performed several ablation tests with our model to validate the effectiveness of each component. Specifically, the following variants were tested: 
(1) without the pre-training process (\textbf{w/o PreT}); 
(2) without the attribute embedding in the encoder (\textbf{w/o AEmb}); 
(3) without the dynamic weight predictor (\textbf{w/o DWP}), i.e., $\lambda_2$ in Eq.~\ref{eq:total_loss} was set to 0 and the outputs from all the attention routes were equally averaged. In order to further demonstrate the effectiveness of the proposed dynamic weighting scheme, we also tested the performance of our model using heuristic weights (\textbf{+ HW}), i.e., $\lambda_2$ in Eq.~\ref{eq:total_loss} was set to 0 and the weight $\alpha$ in Eq.~\ref{eq:mean_att} was set to either 1 or 0 based on the results of the heuristic script during the training.

Moreover, we also tried to generate different responses by setting the weight $\alpha$ in Eq.~\ref{eq:mean_att} to different values in the inference stage. Specifically, we set $\alpha$ to 0 (no persona), 1 (full persona), and the value predicted by the dynamic weight predictor, respectively.

\begin{table}[!t]
\centering
\caption{Automatic evaluation on the random test set.}
\begin{tabular}{lrrrrr}
\toprule
Model             &     Acc.       &    BLEU       &    F1          &  Dist.         &    ppl.        \\ 
\midrule                                                                              
Att+PAB           &     13.99      &    1.61       &     8.60       &   0.130        &    69.30       \\
Trans.            &     7.80       &    3.97       &     12.51      &   0.132        &    43.12       \\  
TTransfo          &     8.80       &    4.06       &     12.63      &   0.169        & \textbf{32.12} \\
TTransfo+P        &     43.05      &    3.44       &     11.28      &   0.158        &    43.78       \\
LConv             &     9.45       &    4.19       &     12.99      &   0.157        &    32.64       \\
LConv+P           &     48.00      &    3.56       &     11.46      &   0.136        &    42.00       \\
\midrule                                                                              
Ours              &     32.80      &    4.18       &     12.52      & \textbf{0.171} &    35.06       \\  
Ours, $\alpha$=1  & \textbf{84.55} &    3.45       &     10.96      &   0.154        &    38.56       \\  
Ours, $\alpha$=0  &     12.90      & \textbf{4.56} & \textbf{13.02} & \textbf{0.171} &    33.71       \\  
w/o PreT          &     27.10      &    3.86       &     11.62      &   0.146        &    48.48       \\
w/o AEmb          &     31.85      &    4.15       &     12.56      &   0.164        &    35.75       \\
w/o DWP           &     30.70      &    4.15       &     12.34      &   0.169        &    34.10       \\
\quad + HW        &     32.55      &    3.50       &     11.90      &   0.151        &    38.52       \\
\bottomrule
\end{tabular}
\label{tab:rand_result}
\end{table}

\subsection{Automatic Evaluation}
\subsubsection{Metrics}
We evaluated the models with the following automatic metrics:
(1) \emph{Persona Accuracy} (\textbf{Acc.}) was computed by feeding the generated responses into the persona classifier together with the target persona, and obtaining the classification accuracy. Higher accuracy values mean the generated responses are more persona consistent. Similar metrics were also used by \citet{zheng2019Personal} and \citet{zhou2018emotional}.
(2) \emph{\textbf{BLEU}}~\citep{papineni-etal-2002-bleu} was used to evaluate how many n-grams (n=1,2) in the generated responses overlap with those in the reference responses.
(3) \emph{\textbf{F1}}~\citep{dinan2019convai2} was calculated based on the character level precision and recall of the generated responses.
(4) \emph{Distinct} (\textbf{Dist.})~\citep{li2015diversity} was used to measure the proportion of unique n-gram in the generated responses (n=1,2).
(5) \emph{Perplexity} (\textbf{ppl.}) was used to measure how the model fits the test data.

\subsubsection{Results}
The performance on the random and biased test set is shown in Table~\ref{tab:rand_result} and Table~\ref{tab:bias_result}, respectively. It can be seen that our model outperforms all the baselines on all the metrics except for the perplexity. This indicates that our proposed model can produce diversified dialogue responses carrying rich persona-related features. We can further observe that:
\textbf{1)} Generating personalized dialogue responses hurts perplexity scores since persona-related words are relative rare and more biased generation of such words will lead to higher perplexity. Though the baseline models fit the test data well (with lower perplexity), they fail to produce more persona-related responses (with lower persona accuracy scores) compared to our model. This observation is in line with the results reported in the ConvAI2 competition~\citep{dinan2019convai2};
\textbf{2)} Ablation tests show that the pre-training stage significantly boost the model performance, and the proposed attribute embedding and dynamic weight predictor also help to improve the performance of our model;
\textbf{3)} The weight $\alpha$ in Eq.~\ref{eq:mean_att} can be used to control the amount of persona-related features to exhibit in the decoding process. Higher $\alpha$ values lead to more persona-consistent responses;
\textbf{4)} Larger performance gaps between our model and the baselines are obtained on the biased test set. This shows that our model can generate more persona-consistent responses in biased contexts.

\begin{table}[!t]
\centering
\caption{Automatic evaluation on the biased test set.}
\begin{tabular}{lrrrrr}
\toprule
Model             &  Acc.          &     BLEU       &    F1          &   Dist.        &    ppl.        \\ 
\midrule                                                                               
Att. + PAB        &   47.60        &     3.08       &     12.50      &   0.133        &    94.38       \\
Trans.            &   34.93        &     7.06       &     15.38      &   0.203        &    85.80       \\  
TTransfo          &   45.87        &     8.68       &     17.39      &   0.260        & \textbf{34.83} \\
TTransfo+P        &   61.61        &     9.10       &     18.41      &   0.257        &    38.07       \\
LConv             &   44.34        &     8.47       &     17.08      &   0.238        &    37.44       \\
LConv+P           &   59.88        &     9.82       &     18.91      &   0.231        &    41.68       \\
\midrule                                                                                          
Ours              &   92.13        &     10.53      &     19.47      &   0.256        &    38.68       \\  
Ours, $\alpha$=1  & \textbf{94.24} & \textbf{11.63} & \textbf{20.51} & \textbf{0.262} &    39.74       \\  
Ours, $\alpha$=0  &   51.44        &     9.00       &     17.44      &   0.249        &    40.89       \\  
w/o PreT          &   71.74        &     9.36       &     18.29      &   0.222        &    95.00       \\
w/o AEmb          &   73.51        &     10.51      &     19.41      &   0.247        &    39.36       \\
w/o DWP           &   73.90        &     10.61      &     19.26      &   0.256        &    37.08       \\
\quad + HW        &   69.87        &     9.01       &     19.81      &   0.232        &    36.37       \\
\bottomrule
\end{tabular}
\label{tab:bias_result}
\end{table}

\begin{figure}[thbp!]
  \centering
  \includegraphics[width=210px]{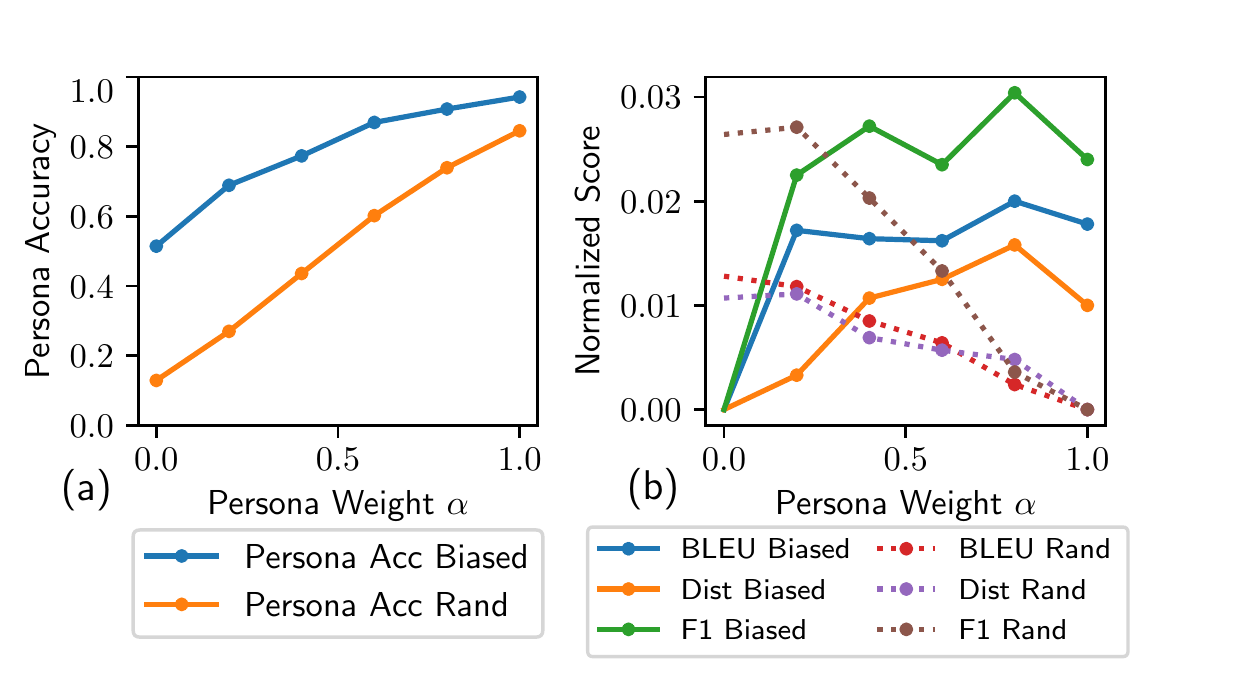}
  \caption{Effects for adjusting the persona weight $\alpha$ in the decoding process. Scores shown on the y-axis of (b) are normalized by subtracting the minimum scores.}
  \label{fig:adjust_w}
\end{figure}

Moreover, the effect of the persona weight $\alpha$ on the generated responses was further evaluated. Specifically, we computed the scores of persona accuracy, BLEU, F1, and distinct corresponding to different $\alpha$ values (Figure~\ref{fig:adjust_w}). It is interesting to observe that:
\textbf{1)} The persona accuracy increases rapidly with $\alpha$ (Figure~\ref{fig:adjust_w}a). This shows that more persona-related features will be incorporated in the decoded responses when $\alpha$ is larger.
\textbf{2)} The scores for BLEU, F1 and distinct on the random test set decrease when $\alpha$ increases (dashed lines in Figure~\ref{fig:adjust_w}b). This is because the dialogues in the random test set are persona-sparse and less overlap between model-produced and human-generated responses will be observed if more persona-related features are incorporated.
\textbf{3)} A clear increasing trend for BLEU, F1 and distinct is observed on the biased test set, but a performance drop is observed when $\alpha$ reaches 1 (solid lines in Figure~\ref{fig:adjust_w}b). This indicates that generating more persona-related responses lead to better performance on the persona-dense contexts, but merely pursuing persona consistency may hurt the performance on other dimensions. This is in line with the manual evaluation results shown in Table~\ref{tab:manual_result}.

\subsection{Manual Evaluation}
\subsubsection{Metrics}
For a given dialogue context and a target persona, we generated responses using all the transformer-based baselines and our model. Three individual annotators were employed to rate the model-generated responses together with the human-generated responses (Gold Resp) from three aspects:
1) \emph{\textbf{Utterance Fluency}}: whether the responses are fluent and could plausibly have been produced by a human; 
2) \emph{\textbf{Persona Consistency}}: whether the responses are consistent with the target persona; 
3) \emph{\textbf{Context Coherency}}: whether the responses are coherent with the dialogue context. 
The rating scale of each measure is of (0, 1, 2), in which rating 0 means worst and 2 best.

\begin{table}[!tp]
\centering
\caption{Manual evaluation on the random and biased test sets.}
\begin{tabular}{p{43pt}p{18pt}p{18pt}p{18pt}p{18pt}p{18pt}p{18pt}}
\toprule
\multirow{3}{*}{Model} & \multicolumn{2}{c}{Utterance} & \multicolumn{2}{c}{Persona} & \multicolumn{2}{c}{Context}           \\
                       & \multicolumn{2}{c}{Fluency} & \multicolumn{2}{c}{Consistency} & \multicolumn{2}{c}{Coherency}       \\
                       \cmidrule{2-7}
                  & Rand           & Biased         & Rand           & Biased         & Rand             & Biased         \\
\midrule                                                 
Trans.            & $1.852$        & $1.810^\dag$   & $0.997^\dag$   & $1.068^\dag$   & $1.428^\dag$     & $1.500$        \\
TTransfo          & $1.832^\dag$   & $1.890$        & $1.015^\dag$   & $1.100^\dag$   & $1.498$          & $1.517$        \\
TTransfo+P        & $1.802^\dag$   & $1.837^\dag$   & $1.125^\dag$   & $1.195^\dag$   & $1.217^\dag$     & $1.483^\dag$   \\
LConv             & $1.863$        & $1.882$        & $1.028^\dag$   & $1.147^\dag$   & $1.490$          & $1.550$        \\
LConv+P           & $1.832^\dag$   & $1.875^\dag$   & $1.093^\dag$   & $1.173^\dag$   & $1.238^\dag$     & $1.478^\dag$   \\
\midrule                                            
Ours              & $1.837^\dag$   & \textbf{1.912} & $1.092^\dag$   & $1.198^\dag$   & $1.487$          & \textbf{1.563} \\
Ours, $\alpha$=1  & $1.835^\dag$   & $1.900$        & \textbf{1.248} & \textbf{1.268} & $1.303^\dag$     & $1.467^\dag$   \\
Ours, $\alpha$=0  & \textbf{1.890} & $1.880^\dag$   & $0.997^\dag$   & $1.085^\dag$   & \textbf{1.535}   & $1.463^\dag$   \\
\midrule                                         
Gold Resp         & $1.928$        & $1.922$        & $1.015$        & $1.423$        & $1.758$          & $1.807$        \\
\bottomrule
\end{tabular}
\label{tab:manual_result}
\raggedright
\small{$\dag$ significant difference with the best result (t-test, $p$-value$<$0.05)}
\end{table}

\subsubsection{Results}
200 dialogue sessions were sampled from each of these two test sets, respectively, and 3.2K responses were generated. The inter-rater annotation agreement was measured using the Fleiss's kappa $\kappa$~\citep{randolph2005free}. Particularly, the $\kappa$ value for \emph{Utterance Fluency}, \emph{Persona Consistency}, and \emph{Context Coherency} was 0.81, 0.70, 0.52, respectively on the random test set, and 0.82, 0.73, 0.49, respectively on the biased test set. This indicates substantial annotation agreement for fluency and persona consistency, and moderate agreement for context coherency.

Table~\ref{tab:manual_result} shows the manual evaluation results. Our model outperforms all the baselines in all the measures. Particularly for persona consistency, our full persona model (i.e., $\alpha$=1) significantly outperforms all the baselines with a large margin. This indicates that our model can generate more persona-consistent responses that are fluent and context-coherent. Further observations also show that:
\textbf{1)} Exhibiting too many persona-related features (i.e., obtaining higher persona consistency) hurts response fluency and context coherency. This is in line with the trade-off between the persona accuracy and perplexity as observed in the automatic evaluation results. Moreover, our dynamic weight predictor provides a better balance between the persona-consistency and context coherency, especially on the biased test set;
\textbf{2)} The persona consistency of our full persona model (i.e., $\alpha$=1) even surpasses the human-generated response on the random test set. This further proves that our model can incorporate richer persona-related features in the generated responses.
\textbf{3)} Although directly fine-tuning on the noisy persona dense data (i.e., TTransfo+P and LConv+P) helps to produce more persona-consistent responses, our model still surpasses these baselines significantly. This verifies the effects of the proposed dynamic weighting scheme. This observation is also in line with the automatic evaluation results shown in Table~\ref{tab:rand_result} and \ref{tab:bias_result}.

\subsection{Case Study}
\begin{figure}[!tp]
  \centering
  \includegraphics[width=240px]{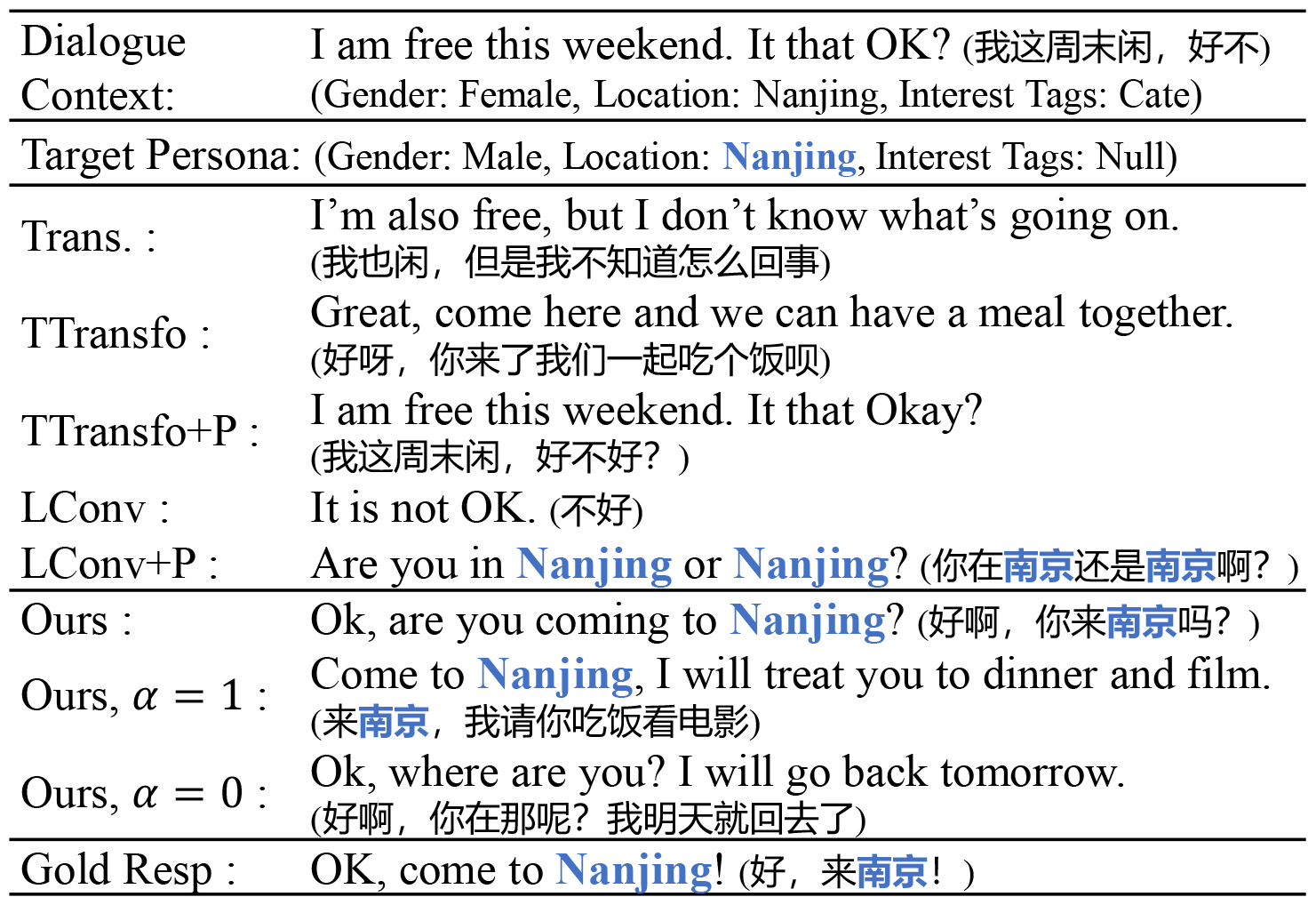}
  \caption{Sample responses generated by baselines and our model.}
  \label{fig:case}
\end{figure}

Figure~\ref{fig:case} shows a sampled case, in which our model can generate coherent responses that reveal rich persona features, while responses produced by the baselines either do not exhibit persona-related features or are not grammatically fluent. This case also shows that the persona weight $\alpha$ can be effectively used to control whether to exhibit persona-related features or not. Specifically, our model with the full persona ($\alpha=1$) reveals the location attribute in the response while our model without persona ($\alpha=0$) does not exhibit persona related features. See the supplementary file for more cases.

\section{Conclusion}
In this paper, we present a pre-training based dialogue generation model that can produce coherent persona-consistent responses conditioned on explicitly represented personas. Our method can effectively utilize persona-sparse dialogue data in the fine-tuning stage. We add attribute embeddings in the encoder to model the persona of each speaker involved in the dialogue context and devise a dynamic weighting scheme in the decoder to balance the amount of persona-related features to exhibit in the decoded responses. Automatic and manual evaluation results show that our model can incorporate richer persona-related features in the generated responses compared to state-of-the-art baselines when the dialogues available at the fine-tuning stage are persona-sparse.

\section{Acknowledgments}
This work was supported by the National Science Foundation of China key project with grant No. 61936010 and regular project with grand No. 61876096, and the National Key R\&D Program of China (Grant No. 2018YFC0830200). We would like to thank Guanyi Chen, Hao Zhou, Chujie Zheng, and Yida Wang for their constructive comments.

\bibliography{ref}
\bibliographystyle{aaai}
\end{document}